\documentclass[runningheads]{llncs}
\usepackage{graphicx}

\usepackage{graphicx}
\usepackage{booktabs}
\usepackage{color}
\usepackage{amssymb}
\usepackage{amsmath}
\usepackage{multirow}
\usepackage{booktabs}

\usepackage{amsfonts}
\usepackage{bm}
\usepackage{import}
\usepackage{cite}

\begin{document}

\title{Multi-Teacher Knowledge Distillation For \\Text Image Machine Translation}

\author{Cong Ma\inst{1,2} \and
Yaping Zhang\inst{1,2}\thanks{Corresponding author.} \and
Mei Tu\inst{4} \and Yang Zhao\inst{1,2} \and Yu Zhou\inst{2,3} \and Chengqing Zong\inst{1,2}}
\authorrunning{C. Ma et al.}

\institute{School of Artificial Intelligence, University of Chinese Academy of Sciences, \\Beijing 100049, P.R. China \and State Key Laboratory of Multimodal Artificial Intelligence Systems (MAIS), Institute of Automation, Chinese Academy of Sciences, Beijing 100190, P.R. China \and Fanyu AI Laboratory, Zhongke Fanyu Technology Co., Ltd, \\Beijing 100190, P.R. China \and Samsung Research China - Beijing (SRC-B)\\
\email{\{cong.ma, yaping.zhang, yang.zhao, yzhou, cqzong\}@nlpr.ia.ac.cn, mei.tu@samsung.com}}

\maketitle              
\vspace{-0.35cm}
\begin{abstract}
Text image machine translation (TIMT) has been widely used in various real-world applications, which translates source language texts in images into another target language sentence. Existing methods on TIMT are mainly divided into two categories: the recognition-then-translation pipeline model and the end-to-end model. However, how to transfer knowledge from the pipeline model into the end-to-end model remains an unsolved problem.
In this paper, we propose a novel Multi-Teacher Knowledge Distillation (MTKD) method to effectively distillate knowledge into the end-to-end TIMT model from the pipeline model.
Specifically, three teachers are utilized to improve the performance of the end-to-end TIMT model. The image encoder in the end-to-end TIMT model is optimized with the knowledge distillation guidance from the recognition teacher encoder, while the sequential encoder and decoder are improved by transferring knowledge from the translation sequential and decoder teacher models. Furthermore, both token and sentence-level knowledge distillations are incorporated to better boost the translation performance.
Extensive experimental results show that our proposed MTKD effectively improves the text image translation performance and outperforms existing end-to-end and pipeline models with fewer parameters and less decoding time, illustrating that MTKD can take advantage of both pipeline and end-to-end models.~\footnote{Our codes are available at: https://github.com/EriCongMa/MTKD\_TIMT}
\end{abstract}
\vspace{-0.65cm}
\section{Introduction}
\vspace{-0.1cm}
Text image machine translation (TIMT) is a cross-modal generation task, which translates source language texts in images into target language sentences. Various real-world applications have been conducted for TIMT, such as digital document translation, scene text translation, handwritten text image translation, and so on. Existing TIMT systems are mainly constructed with a recognition-then-translation pipeline model~\cite{manga_translation, Shekar2021OpticalCR, DBLP:conf/lt4dh/AfliW16, DBLP:journals/ijdar/ChenCN15, DBLP:conf/icdar/DuHSS11}, which first recognizes texts in images by a text image recognition (TIR) model~\cite{DBLP:conf/cvpr/ShiWLYB16, DBLP:journals/pami/ShiBY17, DBLP:conf/iccv/BaekKLPHYOL19, DBLP:conf/prcv/ZhangNLL19, DBLP:journals/tip/ZhangNLL21}, and then generates target language translation with a machine translation (MT) model~\cite{DBLP:conf/nips/SutskeverVL14, DBLP:conf/nips/VaswaniSPUJGKP17, DBLP:conf/ijcai/ZhaoZZZ20, DBLP:conf/coling/ZhaoXZZZZ20}. However, pipeline models have to train and deploy two separate models, leading to parameter redundancy and slow decoding speed. Meanwhile, errors in TIR model are further propagated by MT models, which causes more translation mistakes in the final translation results.

To address the shortcomings of pipeline models, end-to-end TIMT models are proposed with a more efficient architecture~\cite{mansimov-etal-2020-towards}. Although end-to-end models have fewer parameters and faster decoding speed, the end-to-end training data is limited compared with recognition or translation datasets, leading to inadequate training and limited translation performance of end-to-end models. As a result, how to explicitly incorporate external recognition or translation results has been studied by existing research~\cite{DBLP:conf/icpr/ChenYZYL20, DBLP:conf/icpr/MaZTHWZ022}. Furthermore, transfer knowledge from TIR or MT models has been conducted to end-to-end TIMT models through feature transformation~\cite{DBLP:conf/icdar/SuLZ21} and cross-modal mimic framework~\cite{ChenZhuo:TMM}. 

However, sub-modules in end-to-end TIMT models play quite different functions, which need different knowledge from various teacher models. Although existing methods explore to transfer knowledge from external models, how to introduce different knowledge into each sub-modules of the end-to-end TIMT model remains unsolved.

In this paper, we propose a novel multi-teacher knowledge distillation (MTKD) approach for end-to-end TIMT model, which is designed to transfer various types of knowledge into end-to-end TIMT model. Specifically, three sub-modules in end-to-end models are considered to optimize by distilling knowledge from different teacher models.
\begin{itemize}
	\item[$\bullet$] Image encoder aims at extracting features of input images from pixel space to dense feature space, which has a similar function as the TIR image encoder. As a result, TIR image encoder is utilized as the teacher model for image encoder in end-to-end TIMT model to improve the image feature extraction.
	\item[$\bullet$] Sequential encoder in end-to-end TIMT model fuses the local image features into contextual features, which learns advanced semantic information of the sentences in text images. To guide semantic feature learning, MT sequential encoder offers the teacher guidance for TIMT sequential encoder to better map image features into semantic features.
	\item[$\bullet$] Decoder in end-to-end TIMT model generates target translation autoregressively, which has a similar function as the MT decoder. As so, the prediction distribution on target language vocabulary is utilized as the teacher distribution to guide the decoder in end-to-end TIMT generate better prediction distribution.
\end{itemize}

By transferring different knowledge into corresponding sub-modules in end-to-end TIMT model, fine-grained knowledge distillation can better improve the translation quality of end-to-end TIMT models. In summary, our contributions are summarized as:

\begin{itemize}
	\item[$\bullet$] We propose a novel multi-teacher knowledge distillation method for end-to-end TIMT model, which is carefully designed for fine-grained knowledge transferring to various sub-modules in end-to-end TIMT models.
	\item[$\bullet$] Various teacher knowledge distillation provides more improvements compared with single teacher guidance, indicating different sub-modules in end-to-end models need different knowledge information to better adapt corresponding functions.
	\item[$\bullet$] Extensive experimental results show our proposed MTKD method can effectively improve the translation performance of end-to-end TIMT models. Furthermore, MTKD based TIMT model also outperforms pipeline system with fewer parameters and less decoding time.
\end{itemize}

\section{Related Work}

\subsection{Text Image Machine Translation}
Text image machine translation models are mainly divided into pipeline and end-to-end models. Pipeline models deploy text image recognition and machine translation models respectively.
Specifically, the source language text images are first fed into TIR models to obtain the recognized source language sentences. Second, the source language sentences are translated into the target language with the MT model. Various applications have been conducted with the pipeline TIMT architectures. Photos, scene images, document images, and manga pages are taken as the input text images. The TIR model recognizes the source language texts, and the MT model generates target language translation~\cite{DBLP:conf/icpr/WatanabeOKT98, DBLP:journals/pr/ChangCZY09, DBLP:journals/jsw/WongCC11, DBLP:conf/icdar/DuHSS11, DBLP:conf/icassp/YangCZZW02, DBLP:journals/ijdar/ChenCN15, DBLP:conf/lt4dh/AfliW16, manga_translation}. 

End-to-end TIMT models face the problem of end-to-end data scarcity and the performance is limited. To address the problem of data limitation, a multi-task learning method is proposed to incorporate external  datasets~\cite{DBLP:conf/icdar/SuLZ21, DBLP:conf/icpr/ChenYZYL20, DBLP:conf/icpr/MaZTHWZ022}. Feature transformation module is proposed to bridge pre-trained TIR encoder and MT decoder~\cite{DBLP:conf/icdar/SuLZ21}. The hierarchy Cross-Modal Mimic method is proposed to utilize MT model as a teacher model to guide the end-to-end TIMT student model~\cite{ChenZhuo:TMM}.

\subsection{Knowledge Distillation}
Knowledge distillation has been widely used to distillate external knowledge into the student model to improve performance, speed up the training process, and decrease the parameter amounts in teacher models. Specifically, in sequence-to-sequence generation related tasks, token-level and sentence-level knowledge distillation have been proven effective in generation tasks~\cite{DBLP:journals/corr/HintonVD15, DBLP:conf/emnlp/KimR16}. Various tasks have been significantly improved through knowledge distillation method, like bi-lingual neural machine translation~\cite{DBLP:conf/acl/SunWCUSZ20}, multi-lingual translation~\cite{DBLP:conf/iclr/TanRHQZL19}, and speech translation~\cite{DBLP:conf/interspeech/LiuXZHWWZ19}.

To incorporate more knowledge into one student model, multiple teacher models are utilized in some studies to further transfer knowledge into student model. \cite{DBLP:conf/iclr/TanRHQZL19} proposed to use various teacher models in different training mini-batch to make the multilingual NMT model learn various language knowledge. DOPE is designed to incorporate multiple teacher models to guide different subnetworks of the student model to provide fine-grained knowledge like body, hand, and face segmentation information~\cite{DBLP:conf/eccv/WeinzaepfelBC0R20}. 

However, existing methods lack exploration in integrating various knowledge into end-to-end TIMT models. Our proposed multi-teacher knowledge distillation effectively addresses this problem by transferring different knowledge into various sub-modules to meet the corresponding functional characteristics of different modules.

\begin{figure*}[t]
	\centering
	\includegraphics[scale=0.6]{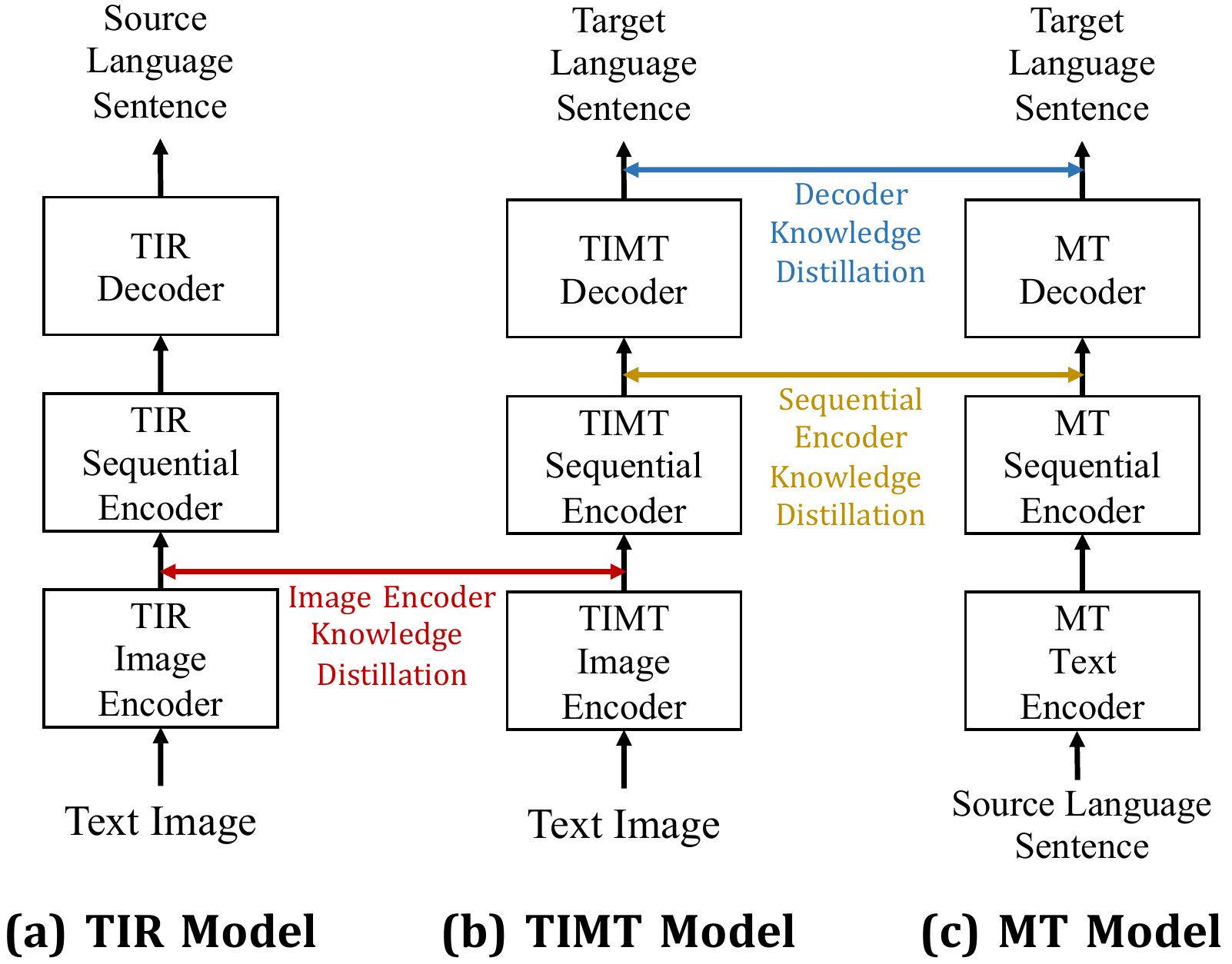}
	\caption{Overall Diagram of (a) Text Image Recognition, (b) Text Image Machine Translation, (c) Machine Translation models and Multi-Teacher Knowledge Distillation.}
	\label{fig_overall_figure}       
	\vspace{-0.35cm}
\end{figure*}

\section{Methodology}
\subsection{Problem Definition}
The end-to-end TIMT model aims at translating source language texts in images into target language sentences. Let $\textbf{I}$ be the source language text image and corresponding target language sentence is $\textbf{Y}$ containing $z$ tokens $\{y^1, y^2, ..., y^z\}$. The training object for the end-to-end TIMT model is to maximize the translation probability:

\begin{equation}
P(\textbf{Y}|\textbf{I};\theta_\text{TIMT}) = \prod_{i=1}^{z}{P(y^{i}|\textbf{I},\textbf{Y}_{<i})}
\end{equation}
where $\textbf{Y}_{<i}$ represents the translation history at the $i$-th decoding step, and $\theta_\text{TIMT}$ denotes the parameters of end-to-end TIMT model.

Specifically, to generate target language translation, end-to-end TIMT model is divided into three sub-modules: image encoder, sequential encoder, and decoder as shown in Figure~\ref{fig_overall_figure} (b). Image encoder $\mathcal{I}$ extracts image features from pixel space and ResNet~\cite{DBLP:conf/cvpr/HeZRS16} is utilized as the image encoder in our work:

\begin{equation}
	F_\mathcal{I} = \mathcal{I}(\textbf{I};\theta_\mathcal{I}) = \text{ResNet}(\textbf{I})
\end{equation}
where $\textbf{I}\in\mathbb{R}^{H \cdot W \cdot C}$ denotes the input text image, and $H, W, C$ represent the height, width, and channel of input image respectively. $F_\mathcal{I}\in\mathbb{R}^{l_\mathcal{I}*c}$ denotes the image feature, and $l_\mathcal{I}, c$ represent length and channel of feature sequence respectively. Generally, image features encoded by convolutional network are $F^{\prime}_\mathcal{I}\in\mathbb{R}^{h \cdot w \cdot c}$, where $h, w, c$ represent the height, width, and channel of feature maps respectively. To meet the requirement of following sequential encoding, feature maps are resized to feature sequence by reducing height and width dimension into feature length: $l_\mathcal{I}=h \cdot w$. Thus, the output of image encoder is a feature sequence containing local information of input text image.

Sequential encoder $\mathcal{S}(\cdot)$ aims at encoding contextual semantic features given local features of input text image. Transformer encoder is utilized as the sequential encoder in this paper:

\begin{equation}
	F_\mathcal{S} = \mathcal{S}(F_\mathcal{I};\theta_\mathcal{S}) = \text{TransformerEncoder}(F_\mathcal{I})
\end{equation}
where $F_\mathcal{S}\in \mathbb{R}^{l_\mathcal{S} \cdot h_\mathcal{S}}$ represents the sequential features that contains contextual semantic information of the whole feature sequence. $l_\mathcal{S}, h_\mathcal{S}$ represent sequence length and hidden dimension of sequential features.

Finally, target language decoder $\mathcal{D}(\cdot)$ generates translation results autoregressively and transformer decoder is utilized in our work:

\begin{equation}
	F_\mathcal{D} = \mathcal{D}(F_\mathcal{S};\theta_\mathcal{D}) = \text{TransformerDecoder}(F_\mathcal{S})
\end{equation}
where $F_\mathcal{D}\in \mathbb{R}^{l_\mathcal{D} \cdot h_\mathcal{D}}$ represents the output of decoder. $l_\mathcal{D}, h_\mathcal{D}$ represent sequence length and hidden dimension of decoder features respectively. The final decoded word $\hat{y}^i_\text{TIMT}$ is calculated by:

\begin{equation}
\label{y-TIMT}
\begin{split}
	\hat{y}^i_\text{TIMT} = \underset{{j}\in [1,|V_\textbf{Y}|]}{\arg\max} \ P(\hat{y}^{i}_j|\textbf{I},\hat{\textbf{Y}}_{<i}), \ \ \ \text{where} \ \ P(\hat{y}^{i}_j|\textbf{I},\hat{\textbf{Y}}_{<i}) \propto W_oF_\mathcal{D}^{i}
\end{split}
\end{equation}
where $P(\hat{y}^{i}_j|\textbf{I},\hat{\textbf{Y}}_{<i})$ denotes the probability that the decoder predicts the $j$-th word $\hat{y}^{i}_j$ in vocabulary at $i$-th decoding step. $W_o\in\mathbb{R}^{|V_\textbf{Y}|\cdot h_\mathcal{D}}$ denotes a linear matrix that maps decoder features into target language words. $|{V}_\textbf{Y}|, h_\mathcal{D}$ represent the size of target language vocabulary and the hidden dimension of decoder respectively. $F_\mathcal{D}^{i}$ means the $i$-th element of decoder feature $F_\mathcal{D}$, which represents the decoder information at position $i$. $\hat{\textbf{Y}}_{<i}$ represents the translation history before $i$-th step. In summary, end-to-end TIMT model utilizes image encoder, sequential encoder, and target language decoder to generate target language translation results word by word.

To optimize the end-to-end TIMT model, the log-likelihood loss function is utilized:
\begin{equation}
\label{L-TIMT}
\begin{aligned}
	\mathcal{L}_\text{TIMT}&=-\sum_{(\textbf{I,Y})\in \textbf{D}_\text{TIMT}}\log P({\textbf{Y}}|\textbf{I})\\
\log P({{\textbf{Y}}}|\textbf{I})&=\sum_{i}^{z}\sum_{j}^{|V_\textbf{Y}|}\mathbb{I}(\hat{y}^i_j=y^i)\log P(\hat{y}^{i}_j|\textbf{I},\hat{\textbf{Y}}_{<i}) 
\end{aligned}
\end{equation}
where  $\mathbb{I}(\hat{y}^i_j=y^i)$ is an indicator function which eques 1 when predicted word $\hat{y}^i$ is the same as the ground-truth $y^i$, otherwise it equals 0. $z$ denotes the sentence length of target language ground-truth. $\textbf{D}_\text{TIMT}$ represents the text image translation training dataset.

\subsection{Architecture of Teacher Models}
Different sub-modules in end-to-end TIMT model play quite different functions and need various knowledge guidance. Image encoder is utilized to extract local visual features from input text images, while a sequential encoder further encodes contextual semantic information from local visual features. Finally, a decoder is designed to generate translation results given sequential features.
To incorporate various knowledge into sub-modules of end-to-end TIMT model, three teacher models are utilized to guide the optimization of image encoder, sequential encoder, and decoder respectively. Specifically, knowledge of extracting text image features is transferred from TIR encoder. MT sequential encoder provides the guidance of contextual semantic feature learning, while MT decoder distillates the target language generation knowledge into TIMT decoder.

\subsubsection{Text Image Recognition Teacher Model.}
Considering image encoder extracts local visual features from input text images, which is consistent between TIMT and TIR tasks, TIR model is incorporated to provide guidance for image feature learning. In this paper, TIR models are also divided into three submodules as end-to-end TIMT model to better understand the information flow between teacher and student models. Similar to TIMT image encoder, TIR image encoder also aims at extracting local visual features of input text images:

\begin{equation}
	F^\text{TIR}_\mathcal{I} = \mathcal{I}^\text{TIR}(\textbf{I};\theta^\text{TIR}_\mathcal{I}) = \text{ResNet}(\textbf{I})
\end{equation}
where $F^\text{TIR}_\mathcal{I}$ denotes the image features encoded by TIR image encoder $\mathcal{I}^\text{TIR}(\cdot)$ and the dimension of $F^\text{TIR}_\mathcal{I}$ is same as the image feature $F_\mathcal{I}$ of end-to-end TIMT model introduced in Section 3.1. $\theta^\text{TIR}_\mathcal{I}$ represents the model parameters of TIR image encoder. The architecture of TIR image encoder is similar to TIMT image encoder, but these two models are trained with different supervised data.

TIR Sequential encoder is also designed to further extract contextual information by considering whole local visual features:
\begin{equation}
	F^\text{TIR}_\mathcal{S} = \mathcal{S}^\text{TIR}(F^\text{TIR}_\mathcal{I};\theta^\text{TIR}_\mathcal{S}) = \text{TransformerEncoder}(F^\text{TIR}_\mathcal{I})
\end{equation}
where $F^\text{TIR}_\mathcal{S}, \mathcal{S}^\text{TIR}(\cdot), \theta^\text{TIR}_\mathcal{S}$ denote TIR sequential features, TIR sequential encoder, and parameters of TIR sequential encoder respectively.

Different from generating target language in TIMT decoder, TIR decoder predicts source language words autoregressively:
\begin{equation}
	F^\text{TIR}_\mathcal{D} = \mathcal{D}^\text{TIR}(F^\text{TIR}_\mathcal{S};\theta^\text{TIR}_\mathcal{D}) = \text{TransformerDecoder}(F^\text{TIR}_\mathcal{S})
\end{equation}
where $F^\text{TIR}_\mathcal{D}, \mathcal{D}^\text{TIR}(\cdot), \theta^\text{TIR}_\mathcal{D}$ denote TIR decoder features, TIR decoder, and parameters of TIR decoder respectively. To further map TIR decoder feature into source language space, a transformation matrix is utilized to transform decoder feature into source language word:
\begin{equation}
\label{x-TIR}
\begin{split}
	\hat{x}^{i}_\text{TIR} = \underset{{j}\in [1,|V_\textbf{X}|]}{\arg\max} \ P(\hat{x}^{i}_j|\textbf{I},\hat{\textbf{X}}_{<i}), \ \ \ \text{where} \ \ P(\hat{x}^{i}_j|\textbf{I},\hat{\textbf{X}}_{<i}) \propto W^\text{TIR}_oF_\mathcal{D}^{\text{TIR}^{i}}
\end{split}
\end{equation}
where $\hat{x}^{i}_j$ represents the $j$-th word in source language vocabulary at decoding position $i$, while $\hat{x}^{i}_\text{TIR}$ represents the final predicted word of decoder at $i$-th decoding step. $W^\text{TIR}_o\in \mathbb{R}^{|V_\textbf{X}|\cdot h^\text{TIR}_\mathcal{D}}$ denotes the transformation matrix from decoder feature space to source language space. $|V_\textbf{X}|, h^\text{TIR}_\mathcal{D}$ represent the size of source language vocabulary and feature dimension of TIR decoder respectively. $F_\mathcal{D}^{\text{TIR}^{i}}$ denotes the TIR decoder feature at position $i$. $\hat{\textbf{X}}_{<i}$ represents the recognition history before $i$-th decoding step.

The overall architecture of TIR and TIMT models is similar, but the supervised data is different. TIR model is trained with recognition data pair $<\textbf{I, X}>$, where $\textbf{X}$ means the source language recognition label of input text image $\textbf{I}$. While TIMT model is trained with text image translation pair $<\textbf{I, Y}>$, where $\textbf{Y}$ means the target language translation of corresponding source language sentence $\textbf{X}$. To optimize the parameters in TIR model, the log-likelihood loss is utilized similar to TIMT optimization:
\begin{equation}
\begin{aligned}
	\mathcal{L}_\text{TIR}&=-\sum_{(\textbf{I,X})\in \textbf{D}_\text{TIR}}\log P({\textbf{X}}|\textbf{I})\\
\log P({\textbf{X}}|\textbf{I})&=\sum_{i}^{z}\sum_{j}^{|V_\textbf{X}|}\mathbb{I}(\hat{x}^i_j=x^i)\log P(\hat{x}^{i}_j|\textbf{I},\hat{\textbf{X}}_{<i}) 
\end{aligned}
\end{equation}
where $\hat{x}^i_j$ denotes the $j$-th word in source language vocabulary at $i$-th decoding step, while $x^i$ represents the ground-truth word at $i$-th decoding step. $z$ denotes the sentence length of ground-truth. $\mathbb{I}(\cdot)$ means the indicator function as introduced in Equation (\ref{L-TIMT}). $\textbf{D}_\text{TIR}$ represents the text image recognition dataset.

\subsubsection{Machine Translation Teacher Model.} Different from cross-modal generation TIR and TIMT models, MT model is a text-to-text transformation network. Thus, the encoder of raw data is quite different from TIR and TIMT models. To obtain text features from source language sentence strings, an embedding layer based text encoder is utilized to map the input words into word embedding:
\begin{equation}
	F^\text{MT}_\mathcal{T} = \mathcal{T}^\text{MT}(\textbf{X};\theta^\text{MT}_\mathcal{T}) = \text{Embedding}(\textbf{X})
\end{equation}
where $F^\text{MT}_\mathcal{T}, \mathcal{T}^\text{MT}(\cdot), \theta^\text{MT}_\mathcal{T}$ represent text features, MT text encoder, and parameters of MT text encoders respectively.

Word embedding only contains single word information rather than global semantic information. To better extract contextual semantic features, MT sequential encoder further encodes contextual information by considering all input words:
\begin{equation}
	F^\text{MT}_\mathcal{S} = \mathcal{S}^\text{MT}(F^\text{MT}_\mathcal{T};\theta^\text{MT}_\mathcal{S}) = \text{TransformerEncoder}(F^\text{MT}_\mathcal{T})
\end{equation}
where $F^\text{MT}_\mathcal{S}, \mathcal{S}^\text{MT}(\cdot), \theta^\text{MT}_\mathcal{S}$ denote MT sequential feature, MT sequential encoder, and parameters of MT sequential encoder respectively. Similar to TIR and TIMT sequential encoder, transformer encoder is utilized to extract contextual semantic features given MT text features.

MT decoder generates target language translation word by word given MT sequential features:
\begin{equation}
	F^\text{MT}_\mathcal{D} = \mathcal{D}^\text{MT}(F^\text{MT}_\mathcal{S};\theta^\text{MT}_\mathcal{D}) = \text{TransformerDecoder}(F^\text{MT}_\mathcal{S})
\end{equation}
where $F^\text{MT}_\mathcal{D}, \mathcal{D}^\text{MT}(\cdot), \theta^\text{MT}_\mathcal{D}$ represent MT decoder features, MT decoder, and parameters of MT decoder respectively. To further map MT decoder features into target language space, a transformation matrix is utilized to calculate the translation probability:

\begin{equation}
\label{y-MT}
\begin{split}
	\hat{y}^{i}_\text{MT} = \underset{{j}\in [1,|V_\textbf{Y}|]}{\arg\max} \ P(\hat{y}^{i}_j|\textbf{X},\hat{\textbf{Y}}_{<i}), \ \ \ \text{where} \ \ P(\hat{y}^{i}_j|\textbf{X},\hat{\textbf{Y}}_{<i}) \propto W^\text{MT}_oF_\mathcal{D}^{\text{MT}^{i}}
\end{split}
\end{equation}
where $\hat{y}^{i}_j$ represents the $j$-th word in target language vocabulary at $i$-th decoding step, while $\hat{y}^{i}_\text{MT}$ represents the final predicted word of target language decoder at decoding position $i$. $\textbf{X}, \hat{\textbf{Y}}_{<i}$ denote source language sentence and translation history before $i$-th decoding step respectively. $W^\text{MT}_o\in\mathbb{R}^{|V_\textbf{Y}|\cdot h^\text{MT}_\mathcal{D}}$ denotes the transformation matrix which maps MT decoder features into target language space. $|V_\textbf{Y}|, h^\text{MT}_\mathcal{D}$ denote the size of target language vocabulary and hidden dimension of MT decoder feature respectively.

\begin{equation}
\begin{aligned}
	\mathcal{L}_\text{MT}&=-\sum_{(\textbf{X,Y})\in \textbf{D}_\text{MT}}\log P(\hat{\textbf{Y}}|\textbf{X})\\
\log P(\hat{\textbf{Y}}|\textbf{X})&=\sum_{i}^{z}\sum_{j}^{|V_\textbf{Y}|}\mathbb{I}(\hat{y}^i_j=y^i)\log P(\hat{y}^{i}_j|\textbf{X},\hat{\textbf{Y}}_{<i}) 
\end{aligned}
\end{equation}
where $\hat{y}^i_j$ denotes the $j$-th word in target language vocabulary at $i$-th decoding step, while $y^i$ represents ground-truth word at $i$-th decoding step. $z$ denotes the sentence length of ground-truth. $\mathbb{I}(\cdot)$ means the indicator function as introduced in Equation (\ref{L-TIMT}). $\textbf{D}_\text{MT}$ represents the text machine translation dataset.

From the comparison of TIR, MT, and TIMT architectures, they have similar and different functions. For example, TIR image encoder and TIMT image encoder have similar structure and functions. All the sequential encoders are similar in architecture and the functions all aim at extracting contextual semantic information. Furthermore, MT decoder and TIMT decoder are both designed to predict target language sentences, which has similar structure and function. As a result, sub-modules of TIMT model with similar architecture and function can as that of TIR or MT models can be improved by multi-teacher knowledge distillation.

\vspace{-0.3cm}
\subsection{Knowledge Distillation from TIR Image Encoder}
TIMT image encoder and TIR image encoder both extract local visual features from input text images. Compared with TIMT task, TIR task has much more training data, thus TIR models can be better optimized to encode image features of text images. To address the data limitation of end-to-end TIMT task, knowledge distillation from TIR image encoder is proposed to transfer text image encoding knowledge into TIMT image encoder. As shown in Figure~\ref{fig_overall_figure}, TIMT image encoder is optimized not only by end-to-end text image translation loss but also by the guidance from TIR image encoder. To align the TIMT image features with TIR image features, both token-level and sentence-level knowledge distillation are incorporated to guide TIMT image encoder to predict similar image features as TIR image features:

\vspace{-0.3cm}
\subsubsection{Token-Level Image Encoder Knowledge Distillation.} TIMT and TIR image features are feature sequences as introduced in Section 3.1. To provide fine-grained guidance information, L2-Norm constraint is utilized to guide TIMT image encoder outputs:
\begin{equation}
	\mathcal{L}_\text{TKD}^{\mathcal{I}}=\frac{1}{B\cdot l_\mathcal{I}}\sum_{j}^{B}\sum_{i}^{l_\mathcal{I}} \Vert F_{\mathcal{I}}^{ij}-F_{\mathcal{I}}^{\text{TIR}^{ij}} \Vert_2
\end{equation}
where $\mathcal{L}_\text{TKD}^{\mathcal{I}}$ denotes the token-level image encoder knowledge distillation loss function. $F_{\mathcal{I}}^{ij}, F_{\mathcal{I}}^{\text{TIR}^{ij}}$ represent TIMT and TIR image features of $j$-th sample at position $i$ respectively. $l_\mathcal{I}$ denotes the length of TIMT image feature sequence, and $l_\mathcal{I}=l_\mathcal{I}^\text{TIR}$ in our experiments, indicating the sequence length of TIMT and TIR image features are the same. $B$ denotes the batch size.

\vspace{-0.3cm}
\subsubsection{Sentence-Level Image Encoder Knowledge Distillation.} To provide sentence-level guidance, both TIMT and TIR global image features are calculated by average pooling:
\begin{equation}
	\mathcal{L}_\text{SKD}^{\mathcal{I}}=\frac{1}{B}\sum_{j}^{B} \Vert \frac{1}{l_\mathcal{I}}\sum_{i}^{l_\mathcal{I}} F_{\mathcal{I}}^{ij}- \frac{1}{l_\mathcal{I}^\text{TIR}}\sum_{i}^{l_\mathcal{I}^\text{TIR}}F_{\mathcal{I}}^{\text{TIR}^{ij}} \Vert_2
\end{equation}
where $\mathcal{L}_\text{SKD}^{\mathcal{I}}$ represents the loss function of sentence-level image encoder knowledge distillation. By calculating the global image features, the optimization of TIMT image encoder is guided by the global alignment between TIMT and TIR image features.

Finally, the token-level and sentence-level image encoder knowledge distillation loss functions are fused to obtain image encoder knowledge distillation loss function $	\mathcal{L}_\text{KD}^{\mathcal{I}}$, which provides multi-granularity knowledge distillation guidance information:

\begin{equation}
	\mathcal{L}_\text{KD}^{\mathcal{I}}=\mathcal{L}_\text{TKD}^{\mathcal{I}}+\mathcal{L}_\text{SKD}^{\mathcal{I}}
\end{equation}

\vspace{-0.3cm}
\subsection{Knowledge Distillation from MT Sequential Encoder}
The sequential encoder is vital to TIMT task, because the contextual semantic features are important for cross-lingual generation. To improve the ability of TIMT sequential encoder, knowledge distillation from MT sequential encoder is incorporated to guide the optimization of TIMT sequential encoder as shown in Figure~\ref{fig_overall_figure}. Similar to image encoder knowledge distillation, sequential encoder knowledge distillation also has token-level and sentence-level knowledge distillations:

\vspace{-0.3cm}
\subsubsection{Token-Level Sequential Encoder Knowledge Distillation.} Similar to the token-level image encoder knowledge distillation, MT sequential features are regarded as the guidance for TIMT sequential features through L2-Norm constraint:
\begin{equation}
	\mathcal{L}_\text{TKD}^{\mathcal{S}}=\frac{1}{B\cdot l_\mathcal{S}}\sum_{j}^{B}\sum_{i}^{l_\mathcal{S}} \Vert F_{\mathcal{S}}^{ij}-F_{\mathcal{S}}^{\text{MT}^{ij}} \Vert_2
\end{equation}
where $\mathcal{L}_\text{TKD}^{\mathcal{S}}$ represents sequential knowledge distillation loss function. $F_{\mathcal{S}}^{ij}, F_{\mathcal{S}}^{\text{MT}^{ij}}$ represent TIMT and MT sequential features of $j$-th sample at position $i$ respectively. $l_\mathcal{S}$ denotes the length of TIMT sequential feature sequence, which is set the same as the length of MT sequential feature sequence $l_\mathcal{S}^\text{MT}$.

\vspace{-0.3cm}
\subsubsection{Sentence-Level Sequential Encoder Knowledge Distillation.} To further provide global guidance of sequential feature learning, the sentence-level sequential encoder knowledge distillation is proposed by performing average pooling on TIMT and MT sequential features: 
\begin{equation}
	\mathcal{L}_\text{SKD}^{\mathcal{S}}=\frac{1}{B}\sum_{j}^{B} \Vert \frac{1}{l_\mathcal{S}}\sum_{i}^{l_\mathcal{S}} F_{\mathcal{S}}^{ij}- \frac{1}{l_\mathcal{S}^\text{MT}}\sum_{i}^{l_\mathcal{S}^\text{MT}}F_{\mathcal{S}}^{\text{MT}^{ij}} \Vert_2
\end{equation}
where $\mathcal{L}_\text{SKD}^{\mathcal{S}}$ denotes the sequential encoder knowledge distillation loss function. The length of TIMT and MT sequential features are the same $(l_\mathcal{S}=l_\mathcal{S}^\text{MT})$ as introduced in token-level sequential encoder knowledge distillation.

Overall sequential encoder knowledge distillation loss function $\mathcal{L}_\text{KD}^{\mathcal{S}}$ is obtained by combining token-level and sentence-level sequential encoder knowledge distillation:
\begin{equation}
	\mathcal{L}_\text{KD}^{\mathcal{S}}=\mathcal{L}_\text{TKD}^{\mathcal{S}}+\mathcal{L}_\text{SKD}^{\mathcal{S}}
\end{equation}

\vspace{-0.3cm}
\subsection{Knowledge Distillation from MT Decoder}
Different from image and sequential encoder knowledge distillation, decoder knowledge distillation is proposed to align the predicted target language vocabulary distribution between TIMT and MT decoders. Token-level decoder knowledge distillation aims at aligning the prediction probability between TIMT and MT decoders at each decoding step, while sentence-level decoder knowledge distillation takes the MT predicted target language sentence as the ground-truth to calculate the decoding loss for the optimization of TIMT model.

\vspace{-0.3cm}
\subsubsection{Token-Level Decoder Knowledge Distillation.} 
As introduced in Equation (\ref{y-TIMT}), TIMT decoder predicts the $j$-th target language word at $i$-th decoding step with the probability of $P(\hat{y}^i_j|\textbf{I},\hat{\textbf{Y}}_{<i}^\text{TIMT})$, while MT decoder generates the $j$-th target language word at $i$-th step with the probability of $P(\hat{y}^i_j|\textbf{X},\hat{\textbf{Y}}_{<i}^\text{MT})$ as in Equation (\ref{y-MT}). To align the decoding distribution, $\textbf{I}$ and $\textbf{X}$ are paired text images and corresponding source language text sentences. $\hat{\textbf{Y}}_{<i}^\text{TIMT}, \hat{\textbf{Y}}_{<i}^\text{MT}$ represent decoding history of TIMT and MT models respectively. The token-level decoder knowledge distillation loss is calculated by updating the vanilla cross-entropy loss:
\begin{equation}
\mathcal{L}^\mathcal{D}_\text{TKD}=-\sum_{i}^{z}\sum_{j}^{|V_\textbf{Y}|}P(\hat{y}^{i}_j|\textbf{X},\hat{\textbf{Y}}_{<i}^\text{MT}) \log P(\hat{y}^{i}_j|\textbf{I},\hat{\textbf{Y}}_{<i}^\text{TIMT}) 
\end{equation}
where $\mathcal{L}^\mathcal{D}_\text{TKD}$ denotes the token-level decoder knowledge distillation loss. By transferring decoding knowledge from MT teacher decoder, the TIMT decoder is guided to have a similar predicted probability of target language words.

\vspace{-0.3cm}
\subsubsection{Sentence-Level Decoder Knowledge Distillation.} To provide sentence-level decoding knowledge distillation, the MT model decoded target language sentences are utilized to replace original ground-truth sentences. Different from token-level decoder knowledge distillation, which is designed to align the decoding probability between TIMT and MT decoders, sentence-level decoder knowledge distillation aims at guiding the TIMT decoder to have similar translation results as MT decoder:
\begin{equation}
\mathcal{L}^\mathcal{D}_\text{SKD}=-\sum_{i}^{z}\sum_{j}^{|V_\textbf{Y}|}\mathbb{I}(\hat{y}^{i}_j=\hat{y}_\text{MT}^{i}) \log P(\hat{y}^{i}_j|\textbf{I},\hat{\textbf{Y}}_{<i}^\text{TIMT}) 
\end{equation}
where $\mathcal{L}^\mathcal{D}_\text{SKD}$ denotes sequence-level decoder knowledge distillation loss function. Different from the vanilla log-likelihood loss function, the ground-truth sentence is replaced as the MT prediction results.  Thus the indicator function $\mathbb{I}(\hat{y}^{i}_j=\hat{y}_\text{MT}^{i})$ equals 1 when the TIMT decoded word $\hat{y}^{i}_j$ is the same as the MT predicted word $\hat{y}_\text{MT}^{i}$. By incorporating both token-level and sentence-level decoder knowledge distillation, the overall loss function of decoder knowledge distillation is formulated as:
\begin{equation}
	\mathcal{L}^\mathcal{D}_\text{KD}=\mathcal{L}^\mathcal{D}_\text{TKD}+\mathcal{L}^\mathcal{D}_\text{SKD}
\end{equation}

The final loss function is the combination of end-to-end text image translation and knowledge distillation loss functions:
\begin{equation}
\begin{aligned}
	\mathcal{L}_\text{ALL}&=(1-\lambda_\text{KD})\mathcal{L}_\text{TIMT}+\lambda_\text{KD}\mathcal{L}_\text{KD} \\
	\mathcal{L}_\text{KD}&=\lambda_\mathcal{I}\mathcal{L}^\mathcal{I}_\text{KD}+\lambda_\mathcal{S}\mathcal{L}^\mathcal{S}_\text{KD}+\lambda_\mathcal{D}\mathcal{L}^\mathcal{D}_\text{KD}
\end{aligned}
\end{equation}
where $\lambda_\text{KD}, \lambda_\mathcal{I}, \lambda_\mathcal{S}, \lambda_\mathcal{D}$ represent the loss weight of overall knowledge distillation, image encoder knowledge distillation, sequential encoder knowledge distillation, and decoder knowledge distillation respectively.

\begin{figure*}[h]
	\centering
	\includegraphics[scale=0.28]{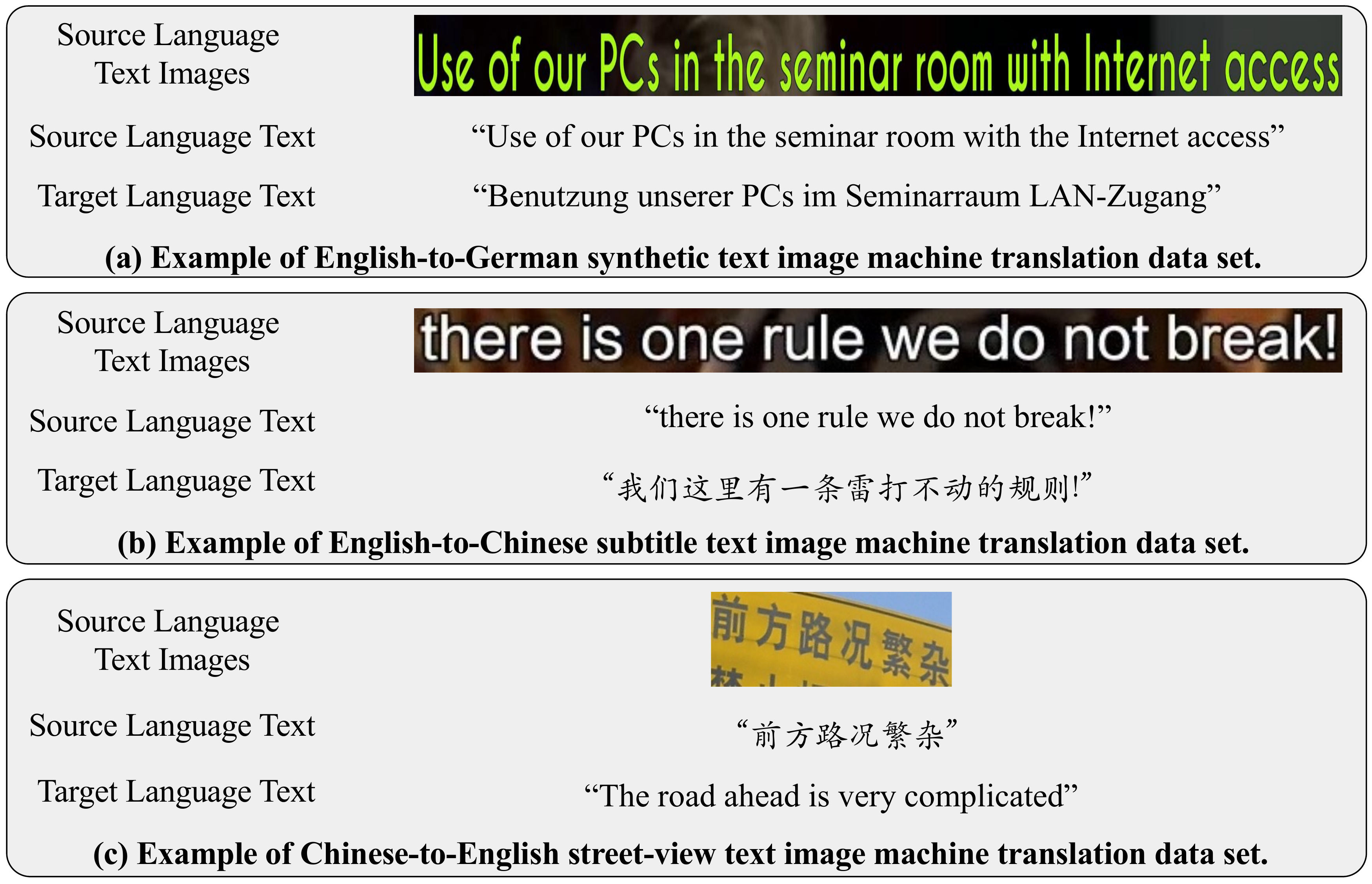}
	\caption{Examples of synthetic, subtitle and street-view text image translation datasets.}
	\label{fig_dataset_example}       
	\vspace{-0.35cm}
\end{figure*}

\vspace{-0.55cm}
\section{Experiments}
\subsection{Datasets}
To train the end-to-end TIMT model, the publicly available dataset released by~\cite{DBLP:conf/icpr/MaZTHWZ022} is utilized in our experiments. As shown in Figure~\ref{fig_dataset_example}, this dataset contains samples from three domains: synthetic, subtitle, and street-view domains. The training and validation samples are all from synthetic domain, while samples in evaluation set are from all three domains. Three translation directions are conducted in this dataset: English-to-Chinese (EnCh), English-to-German (EnDe), and Chinese-to-English (ChEn) translation. There are 1,000,000 training samples, 2,000 validation samples, and 2,000 evaluation samples in synthetic domain. The subtitle test set contains 1,040 samples, while the street-view test set has 1,198 samples. To implement knowledge distillation, triple-aligned samples \{\textbf{source language images, source language texts, target language texts}\} are utilized to transfer the pre-trained knowledge from TIR and MT teacher models into the TIMT student model.

\vspace{-0.3cm}
\subsection{Experimental Setup}
To provide a faire comparison with existing research on end-to-end TIMT task, a similar model architecture as~\cite{DBLP:conf/icpr/MaZTHWZ022} is utilized in our experiment. The TIMT image encoder is composed of TPS Net and Res Net, which extracts the image features from the raw input text images. The TIMT sequential encoder and decoder are 6-layer transformer encoder and 6-layer transformer decoder respectively, which is also the same as~\cite{DBLP:conf/icpr/MaZTHWZ022}. The MT model replaced the TIMT image encoder with an embedding layer based text encoder. The sequential encoder and decoder of the MT model are kept the same as the TIMT model. The preprocessing method and experimental setting are the same as ~\cite{DBLP:conf/icpr/MaZTHWZ022}. For decoding results, sacre-BLEU~\footnote{https://github.com/mjpost/sacrebleu} is calculated to evaluate the translation performance.

\begin{table*}[t]
\centering
\renewcommand{\arraystretch}{1.1}
\setlength{\tabcolsep}{4.7mm}{
\caption{Results of various knowledge distillation combinations on English-to-Chinese translation validation set. TKD and SKD represent using single token-level or sentence-level knowledge distillation loss. TKD+SKD means the fused token-level and sentence-level knowledge distillation are used for knowledge distillation loss function. BLEU Score is utilized to evaluate the translation performance.}
\label{tab_various_kd}
\begin{tabular}{c|ccc|ccc}
\toprule[0.4mm]
No. & $\lambda_\mathcal{I}$ & $\lambda_\mathcal{S}$ & $\lambda_\mathcal{D}$ & TKD & SKD & TKD+SKD \\
\hline
1 & 0 & 0 & 1 & 23.02 & 22.68 & 23.16 \\
2 & 0 & 1 & 0 & 22.63 & 22.44 & 22.85 \\
3 & 0 & 1 & 1 & 23.47 & 23.04 & 23.79 \\
4 & 1 & 0 & 0 & 22.45 & 22.30 & 22.68 \\
5 & 1 & 0 & 1 & 23.28 & 22.95 & 23.52 \\
6 & 1 & 1 & 0 & 23.19 & 22.73 & 23.34 \\
7 & 1 & 1 & 1 & 23.86 & 23.51 & 24.13 \\
\bottomrule[0.4mm]
\end{tabular}}
\vspace{-0.35cm}
\end{table*}

\vspace{-0.3cm}
\subsection{Results of Various Knowledge Distillation}
Table~\ref{tab_various_kd} shows the results of various knowledge distillation (KD) combinations. Line No.1, No.2, and No.4 show the results of single-teacher KD. Single decoder KD (No.1) achieves the best single-teacher performance due to the strong guidance from decoding knowledge. Sequential encoder KD (No.2) outperforms image encoder KD (No.4), indicating semantic knowledge transferring is more important for TIMT task. For bi-teacher KD comparison, sequential encoder and decoder KD combination (No.3) performs well by incorporating semantic and decoding guidance. Finally, triple-teacher KD (No.7) achieves the best performance by transferring image encoder, sequential decoder, and decoder knowledge into end-to-end TIMT model, indicating incorporating accurate knowledge into various sub-modules is vital for performance improvements.

\begin{table*}[t]
\centering
\renewcommand{\arraystretch}{1.1}
\setlength{\tabcolsep}{2.7mm}{
\caption{Comparison of existing end-to-end models with our proposed multi-teacher knowledge distillation (MTKD) method. MTKD utilizes the knowledge distillation setting of line No.7 in Table~\ref{tab_various_kd}.}
\label{E2E_vs_Cascaded_vs_MA}
\begin{tabular}{l|lll|ll|l}
\toprule[0.4mm]
 \multirow{2}{*}{Architecture} & \multicolumn{3}{c|}{Synthetic} & \multicolumn{2}{c|}{Subtitle} & Street \\
  & EnCh & EnDe & ChEn & EnCh & ChEn & ChEn \\
\hline
\multicolumn{7}{c}{Existing End-to-End Models} \\
\hline
 TRBA~\cite{DBLP:conf/iccv/BaekKLPHYOL19} & 9.61 & 7.36 & 4.77 & 12.12 & 5.18 & 0.36 \\
 CLTIR~\cite{DBLP:conf/icpr/ChenYZYL20} & 18.02 & 15.55 & 10.74 & 16.47 & 9.04 & 0.43 \\
 CLTIR+TIR~\cite{DBLP:conf/icpr/ChenYZYL20} & 19.44 & 16.31 & 13.52 & 17.96 & 11.25 & 1.74 \\
 RTNet~\cite{DBLP:conf/icdar/SuLZ21} & 18.91 & 15.82 & 12.54 & 17.63 & 10.63 & 1.07 \\
 RTNet+TIR~\cite{DBLP:conf/icdar/SuLZ21} & 19.63 & 16.78 & 14.01 & 18.82 & 11.50 & 1.93 \\
 \multirow{1}{*}{MTETIMT\cite{DBLP:conf/icpr/MaZTHWZ022}} & 19.25 & 16.27 & 13.16 & 17.73 & 10.79 & 1.69 \\
 MTETIMT+MT\cite{DBLP:conf/icpr/MaZTHWZ022} & {21.96} & {18.84} & {15.62} & {19.17} & {12.11} & {5.84} \\
MHCMM\cite{ChenZhuo:TMM} & 22.08 & 18.97 & 15.66 & 19.24 & 12.12 & 5.87 \\
\hline
\multicolumn{7}{c}{Our Proposed Multi-Teacher Knowledge Distillation Method} \\
\hline
MTKD & \textbf{22.26} & \textbf{19.38} & \textbf{15.84} & \textbf{19.31} & \textbf{12.17} & \textbf{6.08} \\
\bottomrule[0.4mm]
\end{tabular}}
\end{table*}

\vspace{-0.3cm}
\subsection{Comparison with Existing TIMT Methods}
Compared with existing end-to-end TIMT models, MTKD has significant improvements by incorporating various knowledge into sub-modules of TIMT model. Table~\ref{E2E_vs_Cascaded_vs_MA} shows the comparison between MTKD and existing TIMT models. TRBA~\cite{DBLP:conf/iccv/BaekKLPHYOL19} is a vanilla TIR model trained with translation dataset. CLTIR~\cite{DBLP:conf/icpr/ChenYZYL20} proposed to train TIMT model with TIR multi-task learning. RTNet~\cite{DBLP:conf/icdar/SuLZ21} bridges pre-trained TIR and MT models with feature transformer. METIMT~\cite{DBLP:conf/icpr/MaZTHWZ022} trains TIMT model with MT auxiliary task. MHCMM~\cite{ChenZhuo:TMM} is a mimic learning based method by introducing MT teacher for TIMT model. Different from existing research, MTKD incorporates both TIR and MT teachers into TIMT optimization. Meanwhile, various knowledge distillation is utilized to transfer accurate knowledge into sub-modules of TIMT model. Finally, MTKD outperforms the existing best MHCMM model with 0.18 BLEU scores on average. Improvements in all three evaluation domains reveal the good generalization of MTKD.

\begin{table*}[t]
\centering
\renewcommand{\arraystretch}{1.2}
\setlength{\tabcolsep}{2.5mm}{
\caption{Comparison of TIR+MT pipeline method with MTKD method on English-to-Chinese synthetic test set. Model size represents the parameter amount of the model. Decoding time means the time of predicting a sentence and the unit is second. BLEU score is utilized to evaluate the translation performance on valid and test set.}
\label{comparison_of_params_speed}
\begin{tabular}{ccccc}
\toprule[0.4mm]
\multirow{1}{*}{Architecture} & Model Size$\downarrow$ & Decoding Time$\downarrow$ & Valid BLEU$\uparrow$ & Test BLEU$\uparrow$ \\
\hline
Pipeline & 195.1M & 0.33s & 23.52 & 20.46 \\
MTKD & 121.9M & 0.19s & 24.13 & 22.26 \\ 
\bottomrule[0.4mm]
\end{tabular}}

\vspace{-0.45cm}
\end{table*}

\vspace{-0.45cm}
\subsection{Comparison with Pipeline Method}
\vspace{-0.3cm}
Table~\ref{comparison_of_params_speed} shows the comparison of MTKD with the TIR+MT pipeline model. By transferring knowledge into TIMT model, MTKD has better translation performance, which effectively addresses the error propagation problems in pipeline model. With an end-to-end architecture, MTKD has fewer parameters than pipeline model. Meanwhile, MTKD has less decoding time than pipeline model, which is vital in real-world applications.

\begin{figure*}[t]
	\centering
	\vspace{-0.49cm}
	\includegraphics[scale=0.58]{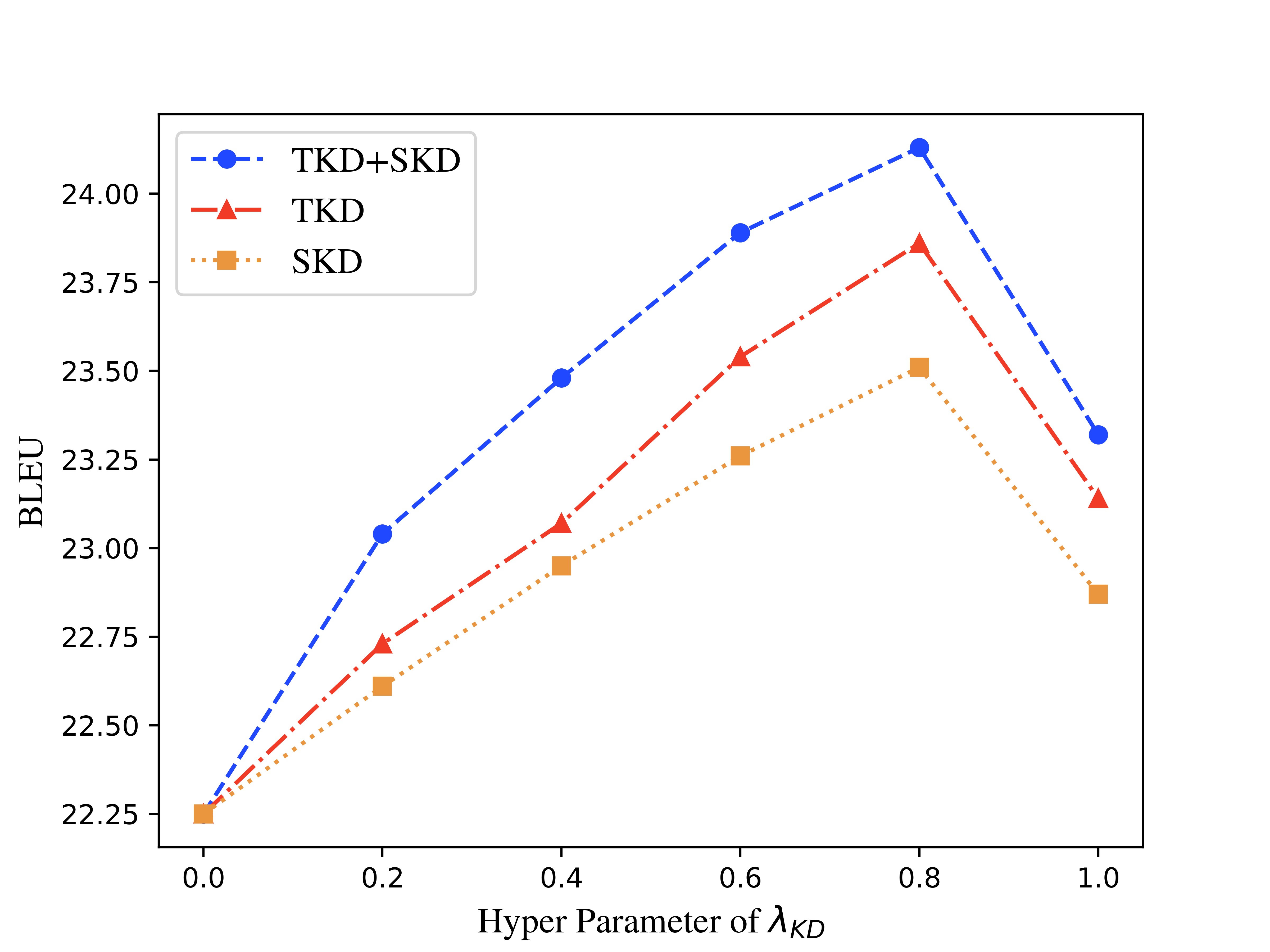}
	\vspace{-0.35cm}
	\caption{Hyper-parameter analysis on the loss weight of knowledge distillation.}
	\label{fig_hyper_para}       
	\vspace{-0.45cm}
\end{figure*}

\vspace{-0.3cm}
\subsection{Analysis of Hyper-parameter}
\vspace{-0.15cm}
The loss weight of knowledge distillation is a key hyper-parameter to balance the end-to-end TIMT loss and knowledge distillation losses. When $\lambda_\text{KD}=0$, the model is only optimized with end-to-end loss function and the performance is limited due to the end-to-end data scarcity and the difficulty of TIMT task. By incorporating KD loss, the performance is getting better and the optimal value for $\lambda_\text{KD}$ is 0.8. When $\lambda_\text{KD}=1$, the performance drops a bit, indicating end-to-end TIMT loss by guiding the model learns to predict as the ground-truth is also important for TIMT task.

\vspace{-0.45cm}
\section{Conclusion}
\vspace{-0.3cm}
In this paper, we propose a novel multi-teacher knowledge distillation (MTKD) method for end-to-end text image machine translation task. Three pre-trained teacher models are utilized to provide accurate knowledge for corresponding sub-modules in end-to-end TIMT model. By transferring various knowledge into sub-modules of TIMT model, the translation performance is significantly improved compared with existing methods. Meanwhile, token-level and sentence-level knowledge distillation are complementary for knowledge transferring, indicating that multi-granularity knowledge distillation is vital for TIMT improvements. Furthermore, MTKD based TIMT model outperforms pipeline models with a smaller model size and less decoding time, which has the advantages of both end-to-end and pipeline models. In the future, we will explore to transfer more knowledge into end-to-end TIMT model to further improve the translation performance.

\vspace{-0.55cm}
\section*{Acknowledgement}
\vspace{-0.175cm}
This work has been supported by the National Natural Science Foundation of China (NSFC) grants 62106265.

\bibliographystyle{splncs04}
\bibliography{refs}

\end{document}